# ADAPTIVE WAVELET BASED IDENTIFICATION AND EXTRACTION OF PQRST COMBINATION IN RANDOMLY STRETCHING ECG SEQUENCE

*T.R Gopalakrishnan Nair*  
RIIC, DSI  
Bangalore, India

*Geetha A P*  
RIIC, DSI  
Bangalore, India

*Asharani M*  
Dept.of ECE, JNTUHEC  
Hyderabad, India

## ABSTRACT

Cardiovascular system study using ECG signals have evolved tremendously in the domain of electronics and signal processing. However, there are certain floating challenges unresolved in the analysis and detection of abnormal performances of cardiovascular system. As the medical field is moving towards more automated and intelligent systems, wrong detection or wrong interpretations of ECG waveform of abnormal conditions can be quite fatal. Since the PQRST signals vary their positions randomly, the process of locating, identifying and classifying each feature can be cumbersome and it is prone to errors. Here we present an automated scheme using adaptive wavelet to detect prominent R-peak with extreme accuracy and algorithmically tag and mark the coexisting peaks P, Q, S, and T with almost same accuracy. The adaptive wavelet approach used in this scheme is capable of detecting R-peak in ECG with 99.99% accuracy along with the rest of the waveforms.

## 1. INTRODUCTION

One of the non-invasive methods to register the electrical activity of heart is ECG. The changes in electrical potential during depolarization and repolarisation of the myocardial fibres are recorded by electrodes positioned on the surface of the body. The normal ECG recordings have a number of different morphologies depending of the patient, type of the lead used for recording etc. So it is hard to build a universal tool for automatic ECG analysis. Computational intelligence can bring forward a number of conceptually and computationally appealing methods for ECG signal processing, classification and interpretation. The non-invasive method of recording and analyzing the ECG signal has made it popular as a routine part of any complete medical evaluation.

The normal clinical features of the electrocardiogram, which include wave amplitudes and inter-wave timings, are shown in Fig 1. The iso-electric line on ECG is a horizontal line when there is no electrical activity in heart. The first deflection shown in the ECG is the P-wave. It results from depolarization of the atria. QRS complex corresponds to the ventricular depolarization. The T-wave represents ventricular repolarisation. In few individuals a U- wave may be visible after T-wave which is caused by the after-potentials that are probably generated by mechanical–electric feedback [1]. The amplitude and relative position of different waves P-Q-R-S-T give valuable information about the functioning of heart. There are various methods adapted for the detection and identification of waveform features.

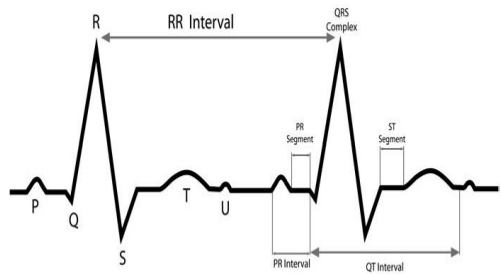

Fig. 1 A normal ECG waveform

Automatic detection of ECG features depends mainly on the accurate detection of R- peak. R-peak detection using slope- amplitude analysis [2], digital filters [3-6], difference operation method [7] and transformed domains [8] were the first few developed methods. Artificial Neural Networks [9, 10], Genetic Algorithm [11], Hidden Markov Model [12] and Support Vector Machines [13], Shannon energy envelope (SEE) estimator[14] were also used for the QRS detection. Wavelet transform has emerged over recent years as a powerful tool for de-noising and detection of QRS complex.

For the detection of characteristic points of ECG waveforms, Li et al. had introduced an algorithm based on discrete wavelet transform (DWT) [15]. Identifying the singularity points using Lipschitz exponent at different scales was the adapted method. Kadambe et al. also used the DWT analysis using Spline wavelet as they are insensitive to non-stationarity in the QRS complex and robust to noise [16]. Mahmoodabadi et al. analysed ECG signals from Modified Lead II using two filters D4 and D6 and compared their performance. In order to delineate the ECG waveform Martínez et al. used a quadratic spline wavelet with adaptive threshold levels [17]. Sivanarayana and Reddy have used bi-orthogonal wavelet transform for ECG parameters estimation [18]. Virgilio et al. detected characteristic points by comparing the coefficients of the discrete WT on selected scales against fixed thresholds [19]. Romero et al. had used Continuous Wavelet Modulus Maxima for R-wave detection [20]. A relative performance analysis about the different wavelet transform analysis method for the QRS detection was presented by Senhadji et al. [21].

All the attempts hitherto undoubtedly were a step forward in accurate detection of R peak. Many of the methods used 4 to 6 scales of resolution to find the characteristic points. Algorithm was tested with one database and obtained accuracy varying from 95 % to 995. In our study we have used a QRS-pattern adaptive wavelet for continuous wavelet transform analysis of ECG signals to bring out high accuracy R-peak detection. Once the R-peaks are marked other characteristic points P-Q-S-T are identified accurately in time domain analysis.

Continuous wavelet transform (CWT) is preferable over DWT for signal analysis, feature extraction and detection tasks, for it provide a description that is truly shift invariant. CWT performs a correlation analysis, so that it gives maximum when the input signal resemble to the mother wavelet. The continuous wavelet transform (CWT) of a signal x (t) is defined as

$$W(a,b) = \frac{1}{\sqrt{a}} \int_{-\infty}^{\infty} x(t) \Psi^*\left(\frac{t-b}{a}\right) dt \quad (1)$$

where $\Psi^*(t)$ is the complex conjugate of the analysing wavelet function ψ(t), parameters a and b is the dilation and location parameter of the wavelet respectively. Here x (t) is decomposed into a set of basis functions Ψ(t) called the wavelets, along the new dimensions, scale and translation. The most important properties of wavelets are the admissibility and the regularity conditions. The admissibility condition can be used to first analyze and then reconstruct a signal without loss of information. Regularity conditions state that the wavelet function should have some smoothness and concentration in both time and frequency domains. In other words it should have finite energy.

Wavelet transforms can comprise an infinite set of possible basis functions. Primary wavelets or the adaptive wavelets are used as basis functions. Most of the ECG analysis were carried out using primary wavelets such as spline, Mexican hat, Haar, Daubechies etc. As CWT performs a correlation study, using a pattern adaptive wavelet can give a better analysis. Even in the presence of noise an adaptive wavelet can give a better correlation. For our study we developed an adaptive wavelet matching to the QRS complex of a typical ECG waveform. This paper brings out an accurate method to identify ECG characteristic points using adaptive wavelet.

## 2. RESEARCH BACKGROUND

Any clinical analysis of ECG waveform starts from identifying the QRS complex, its amplitude and width as well as its regularity. This will be followed by P-wave and T- wave analysis, and the analysis of various intervals, P- R, R-R, S-T and Q-T segments. The primary objective of any digital analysis of ECG wave is to locate the R-peak accurately. ECG waveforms are usually contaminated by noise and artifacts such as power line interference, contact noise, patient– electrode motion artifacts, electromyographic noise, baseline drift etc. [1]. Accurate identification of R-peak and other characteristic points still remain a challenging task because of the presence of noise and the varying morphologies of ECG waves.

Time-frequency analysis techniques of WT have been applied for ECG analysis, to take advantage of the nonstationary nature of the cardiac cycle. But most of the analysis of ECG waves were mainly carried out using DWT. Romero et al. used CWT using Mexican hat for ECG signal time-frequency analysis. In CWT as there is no discretization of scales, it can provide a better resolution. Using a high resolution in wavelet space allows individual maxima to be followed accurately across scales.

In this paper, we have overcome the difficulties experienced in the earlier works like low detection sensitivity in the presence of noise, computational complexity associated with the requirement to calculate wavelet transform over many scales etc. We used CWT analysis of the ECG signal using an adaptive wavelet designed for QRS complex detection. An adaptive wavelet is the best basis function designed for a given signal representation. There are several methods to construct pattern-adaptive wavelets. Bi-orthogonal method [22], Projection based methods [23], statistical method [24], Lifting scheme [25] etc. Misiti et al. have used the method of least square optimization for generating pattern adaptive wavelet [26]. The principle for designing a new wavelet for CWT is to approximate a given pattern, using least squares optimization under constraints, leading to an admissible wavelet compatible for the pattern detection. In our research work, we have used this method with a polynomial approximation of order 4 to generate the adaptive wavelet. Using adaptive wavelet R-peaks were detected accurately.

Many of the previous studies were mainly focused on the QRS complex detection. Abed et al. has developed an algorithm to detect P, QRS and T waves. Authors have used Haar wavelet for QRS detection and later DB2 wavelet was used for P and T detection [27]. Chouhan and Mehta employed a modified definition of slope of ECG signal, as the feature for detection of ECG wave components [28]. For myocardial ischemia analysis Ranjit et al. considered WT up to four scales and the scale 2 to 4 was used to locate T- and P-waves using maxima minima method [29]. In our study once the R- peaks were detected using adaptive wavelet, Q-S-P-T points were identified with high accuracy in time domain analysis.

## 3. RESEARCH METHODOLOGY

After analysing various ECG signals from the database [30] a typical bipolar QRS waveform was selected. Using the method of least square optimization, a pattern adaptive wavelet was generated [28]. The adaptive wavelet is shown in Fig. 3. A wavelet was adapted for the pattern using Matlab® Wavelet toolbox command 'pat2cwav', i.e. 'pattern to continuous wave'. The function gives an approximation to the given pattern in the interval [0 1] by least squares fitting, a projection within the space of functions orthogonal to constants. This adaptive wavelet is used for the CWT analysis of the ECG signals.

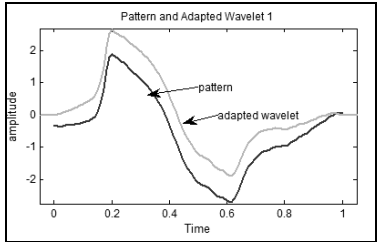

Fig. 2 Adaptive wavelet

ECG signals required for analysis were collected from global data source where annotated ECG signals are available [30]. MIT-BIH Arrhythmia and PTB diagnostic database were used for the testing of algorithm.

3.1 Baseline Drift Removal

Any ECG signal analysis requires a pre-processing to remove the noise artifacts. For computerized detection of QRS complexes based on threshold detection the main affecting noise factor will be baseline drift. The frequency content of the baseline wander is usually in a range well below 0.5Hz. This baseline drift can be eliminated using median filters (200-ms and 600-ms) [30]. The original ECG signal was processed with a median filter of 200-ms width to filter QRS complexes and P waves. The resulting signal was then processed with a median filter of 600-ms width to remove T waves. The signal resulting from the second filter operation contained the baseline of the ECG signal, which was then subtracted from the original signal to produce the baseline corrected ECG signal. The baseline corrected signal for the ECG signal 108 V1 lead is shown in Fig 3. The intermediate results of unravelling QRS & P-wave as well as T-waves are also shown in fig 3.

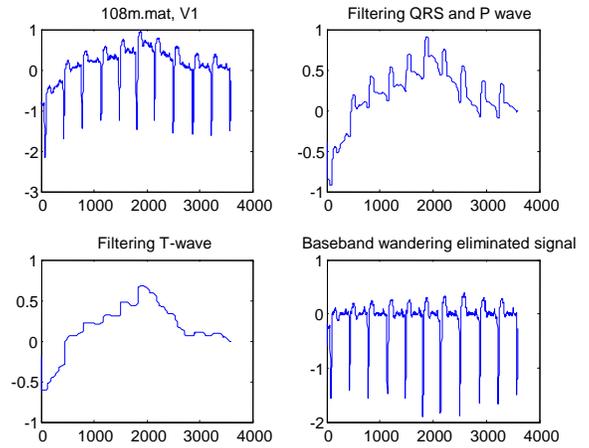

Fig. 3 Base Band Eliminated Signal

3.2 De-noising the ECG Wave

Another factor which can affect the peak detection process is the low frequency noise. Detection of R peak will not be affected by the presence of noise as the peak amplitude is high. But the P, Q, S and T wave detection will be affected by the noise as these waves are of low amplitude. Stationary wavelet transform with DB1 is used for the noise removal. De-noised signal is shown in fig. 4.

*3.3 Suppressing T-Wave*

Since high T-wave can lead to false detection it is better to suppress the T-wave before the QRS detection. From the de-noised signal, T-wave can be suppressed using the median filter of 600ms width. A suppressed T-wave waveform is shown in Fig. 5 for the ECG signal 102, V2 lead.

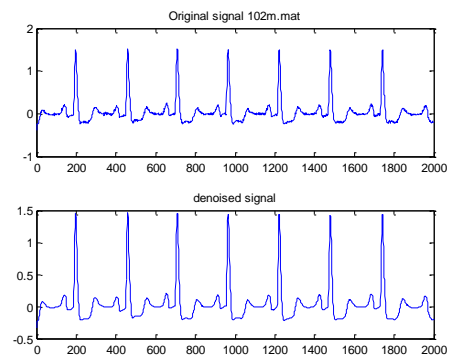

Fig. 4 De-noised signal

Before finding out the R-peaks using adaptive wavelet the signal is squared. Squaring of the signal allows us to use the same adaptive wavelet for all waveforms for all the different leads from which the ECG signal is being taken. That means for analysing "bipolar signal", "only negative amplitude signal" OR "only positive amplitude signal", the same adaptive signal can be used. It also enhances the R-peak by reducing the P-wave effect as shown in Fig 5. The high frequency components in the signal related to the QRS complex are further enhanced. Squaring of signal also helps in analysing waveforms of very low signal to noise ratio. Following are the major steps involved in finding the R-peaks.

1. Calculate continuous wavelet transform of the given ECG signal up to two levels using the adaptive wavelet.
2. Detect a series of maximum- minimum pair in the first and second level.
3. Remove those maximum - minimum pairs whose absolute values are less than the threshold. Threshold value is selected as the 30% of the maximum of CWT coefficients as shown in Fig. 5. Statistical and heuristic estimation methods were used to arrive at the threshold value. Threshold value less than 30% results in higher false detection, whereas threshold value greater than 30% results in more number of misses in peak detection.
4. Detect the zero - crossing point between a pair of maximum and minimum points.
5. Decrease the false detections by removing peaks which are occurring within less than 120ms.
6. Verify second level coefficients in a similar manner and compare the R-peak positions to avoid false detection.
7. Previously rejected events are re-evaluated using a reduced threshold when a significant time has passed without finding a QRS complex. For ECG signals with R_R peak variability, reducing the time limit for less than120ms may be required.
8. From R-peak point, moving to the left and right to the extent of 15% (of RR interval) find the first trough/peak (if R-peak is positive, find the trough, if R- peak is negative find the peak) to locate the Q and S respectively. Q and S points should be a point after crossing the iso-electric line. Positions of P and T waves come approximately within 40% of RR interval from Q and S points. From Q moving to the left find the first maximum point to fix P- point and from S moving to the right the first peak gives T point. If there is a negative occurrence of T wave it is detected by comparing the negative and positive peak within the 40% RR interval. The Fig. 6 shows an ECG wave with all these points marked.

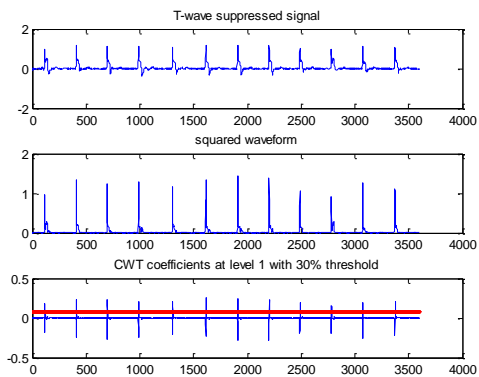

Fig. 5 Intermediate processed ECG signal

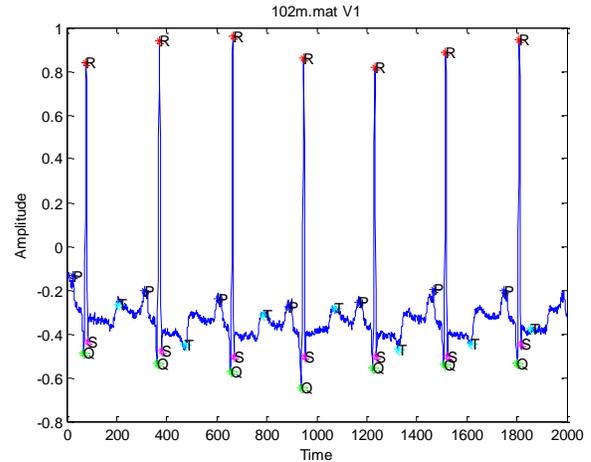

Fig.6 PQRST marked ECG signal

## 4. RESULT

The proposed algorithm was tested on two different sets of data, MIT-BIH arrhythmia database and PTB diagnostic database. For MIT-BIH database the sampling frequency was 360Hz with 2- channels and for PTB database the sampling frequency was 1 KHz with 16 channels. In both databases it was possible to detect all the characteristic peaks PQRST. For few signals (like 101 and 106 waveform from MIT-BIH), the noise level was very high, where only R detection was possible and other peaks were not distinguishable. Inverted T –waves were also detected with high accuracy. Cases of atrial fibrillation were not considered for testing as only R –peaks are present in the signal. Special cases of missed or merged P-waves were analysed separately by taking the relative positions of P, T and R-waves to avoid false P-detection.

## 5. CONCLUSION

Adaptive wavelet approach has provided an R-peak detection accuracy of 99.9% by making use of just 2 levels of resolution. Same accuracy of detection was obtained for both databases having different sampling rates. With the effective baseline correction and noise removal, other characteristic points viz P-Q-S-T were also detected with high accuracy.

The algorithm has not incorporated the automatic detection of atrial fibrillation (AF). The research can be extended to incorporate the automatic detection of AF. The research uses currently available databases, analysing real time signals and providing real time results will be the next challenge.

## 5. REFERENCES

[1] Gari D. Clifford, Francisco Azuaje, Patrick E. McSharry, Advanced Methods and Tools for ECG Data Analysis

[2] Jiapu Pan and Willis J. Tompkins, "A Real-Time QRS Detection Algorithm, IEEE Transactions on Biomedical Engineering," Vol. BME-32, No. 3, March 1985

[3] M.L. Ahlstrom and W.J. Tompkins, "Digital filters for real time ECG signal processing using microprocessors," IEEE Trans. Biomed. Eng., vol. BME-32, no. 9, pp. 708-713, Sept. 1985.

[4] G. Tremblay and A.R. LeBlanc, "Near-optimal signal preprocessor for positive cardiac arrhythmia identification," IEEE Trans. on Biomed. Engg., vol.32, no.2, pp.141-151, Feb. 1985.

[5] Q. Xue, Y.H. Hu and W.J. Tompkins, "Neural-network based adaptive matched filtering for QRS detection," IEEE Trans. Biomed. Eng., vol. 39, no. 4, pp. 317-329, April 1992.

[6] S. Suppappola and Sun Ying, "Nonlinear transforms of ECG signals for digital QRS detection: a quantitative analysis," IEEE Trans. Biomed. Eng., vol. 41, pp. 397-400, April 1994.

[7] Yun-Chi Yeha,c, Wen-June Wang, " QRS complexes detection for ECG signal: The Difference Operation Method," computer methods and programs in biomedicine ( 2008 ) 245

[8] I.S.N. Murthy and G.S.S.D. Prasad, "Analysis of ECG from pole-zero models," IEEE Trans. on Biomed. Engg., vol.39,no.7, pp.741-751, July 1992

[9] Hu YH, Tompkins WJ, Urrusti JL, Afonso VX., "Applications of artificial neural networks for ECG signal detection and classification.," J Electrocardiology. 1993; Suppl:66-73.

[10] Y. Suzuki, "Self-organizing QRS-wave recognition in ECG using neural networks," IEEE Trans. Neural Networks, vol. 6, pp. 1469-1477, 1995.

[11] R. Poli, S. Cagnoni and G. Valli, "Genetic design of optimum linear and nonlinear QRS detectors," IEEE Trans. Biomed. Eng., vol. 42, no. 11, pp. 1137-1141, Nov. 1995.

[12] D.A. Coast, R.M. Stern, G.G. Cano and S.A. Briller, "An Approach to Cardiac Arrhythmia Analysis Using Hidden Markov Models," IEEE Transactions on Biomedical Engineering, vol. 37, no. 9, pp. 826-836, Sept. 1990.

[13] Mehta SS, Lingayat NS., "Detection of QRS complexes in electrocardiogram using support vector machine," J Med Eng Technol. 2008 May-June;32(3):206-15.

[14] M.Sabarimalai, Manikandana, K.P.Soman, "A novel method for detecting R-peaks in electrocardiogram(ECG) signal," Biomedical Signal Processing and Control (2011)

[15] Cuiwei Li, Chongxun Zheng, and Changfeng Tai, "Detection of ECG Characteristic Points using Wavelet Transforms," IEEE Transactions on Biomedical Engineering, Vol. 42, No. 1, pp. 21-28, 1995.

[16] Kadambe S, Murray R and Boudreaux-Bartels G F, "Wavelet transform-based QRS complex detector," IEEE Trans. Biomed. Eng. 46 838–48, 1999

[17] Martín´ez JP, Almeida R, Olmos S "A Wavelet-Based ECG Delineator Evaluation on Standard Databases," IEEE Trans Biomed Eng 2004; 51(4):570–581.

[18] N. Sivannarayana, D. C. Reddy, "Biorthogonal wavelet transforms ELSEVIER 1999; 21:167-174.

[19] V. Di-Virgilio, C. Francaiancia, S. Lino, and S. Cerutti, "ECG fiducial points detection through wavelet transform," in 1995 IEEE Eng. Med. Biol. 17th Ann. Conf. 21st Canadian Med. Biol. Eng. Conf., Montreal, Quebec, Canada, 1997, pp. 1051-1052.

[20] Romero Legarreta, PS Addison, N Grubb,GR Clegg, CE Robertson, KAA Fox , JN Watson, "R-wave Detection Using Continuous Wavelet Modulus Maxima," IEEE Computers in Cardiology 2003;30:565−568.

[21] Senhadji L, Carrault G, Bellanger J J and Passariello G "Comparing wavelet transforms for recognizing cardiac patterns," IEEE Trans. Med. Biol. 13 167–73, 1995

[22] Joseph O. Chapa and Raghuveer M. Rao "Algorithms for Designing Wavelets to Match a Specified Signal," IEEE transactions on signal processing, vol. 48, no. 12, December 2000

[23] Akram Aldroubi, Patrice Abry, and Michael Unser "Construction of Biorthogonal Wavelets Starting from Any Two Multiresolutions," IEEE transactions on signal processing, vol. 46, no. 4, April 1998

[24] Gupta, A. Joshi, S.D. Prasad, S. "A new approach for estimation of statistically matched wavelet," IEEE Transactions on Signal Processing,vol.53 no.5 May 2005

[25] Wim Sweldens , Peter schroder "Building Your Own Wavelets at Home," Technical report Industrial Mathematics initiative. Department of Mathematics, University of South Carolina 1995.

[26] Misiti, Y. Misiti, G. Oppenheim, J.M. Poggi, Hermes, "Les ondelettes et leurs applications," M, 2003.

[27] Abed Al Raoof Bsoul, Soo-Yeon Ji, Kevin Ward, and Kayvan Najarian, "Detection of P, QRS, and T Components of ECG Using Wavelet Transformation," Complex Medical Engineering, 2009 CME, ICME, International

[28] V.S. Chouhan† and S.S. Mehta, "Threshold-based Detection of P and T-wave in ECG using New Feature Signal," IJCSNS International Journal of Computer Science and Network security, Vol.8 No.2, February 2008

[29] P. Ranjith, P.C. Baby, P. Joseph, "ECG analysis using wavelet transform: Application to Myocardial Ischemia Detection," IRBM, Elsevier Volume 24, Issue 1, February 2003, Pages 44–47.

[30] www.physionet.org/physiobank